# Mapping Violence: Developing an Extensive Framework to Build a Bangla Sectarian Expression Dataset from Social Media Interactions

**Nazia Tasnim**[*]
Boston University

**Sujan Sen Gupta**[*]
University of Dhaka

**Md. Istiak Hossain Shihab**
Oregon State University

**Fatiha Islam Juee**
University of Dhaka

**Arunima Tahsin**
University of Dhaka

**Pritom Ghum**
ICDDR,b

**Kanij Fatema**
Bengali.AI

**Marshia Haque**
Bengali.AI

**Wasema Farzana**
BRAC University

**Prionti Nasir**
Microsoft

**Ashique KhudaBukhsh**
RIT

**Farig Sadeque**[†]
BRAC University, Bengali.AI

**Asif Sushmit**[†]
RPI, Bengali.AI

## Abstract

Communal violence in online forums has become extremely prevalent in South Asia, where many communities of different cultures coexist and share resources. These societies exhibit a phenomenon characterized by strong bonds within their own groups and animosity towards others, leading to conflicts that frequently escalate into violent confrontations. To address this issue, we have developed the first comprehensive framework for the automatic detection of communal violence markers in online Bangla content accompanying the largest collection (13K raw sentences) of social media interactions that fall under the definition of four major violence class and their 16 coarse expressions. Our workflow introduces a 7-step expert annotation process incorporating insights from social scientists, linguists, and psychologists.

By presenting data statistics and benchmarking performance using this dataset[1], we have determined that, aside from the category of *Non-communal violence*, *Religio-communal violence* is particularly pervasive in Bangla text. Moreover, we have substantiated the effectiveness of fine-tuning language models in identifying violent comments by conducting preliminary benchmarking on the state-of-the-art Bangla deep learning model.

## 1 Introduction

Internet has revolutionized the way people communicate and connect with each other. However, along with its advantages, it has also instigated the spread of hate speech, harassment, and other forms of online violence. Apparent anonymity provided by the internet can encourage individuals to express extreme opinions and use language they would not use in face-to-face interactions- a phenomenon referred to as Dis-inhibition Effect (Suler, 2004). Social media algorithms and echo chambers can further exacerbate the problem by amplifying extremist views, making it easier for hateful ideas to spread and gain momentum (Benkler et al., 2018; Pariser, 2011; Sunstein, 2018). In addition, the spread of false information online can fuel conspiracy theories and other forms of hate speech (Mathew et al., 2019), leading to increased tension and potential violence. The internet can also foster polarization and tribalism (Van Alstyne and Brynjolfsson, 2005), with individuals strongly identifying with their social group and viewing other groups as adversaries. It fuels extremist groups to mobilize and spread their message, often targeting vulnerable individuals or communities (Yang et al., 2022).

Acknowledging online violence as a *spectrum* allows for a more nuanced approach to addressing harmful online issues. It recognizes that different forms of violence can have varying levels of severity and that some forms are more readily visible than others. This ranges from casual animosities to descriptive dehumanization, supporting discrimination, disseminating propaganda, and directly inflicting harm that may lead to mental or physical impairment. This corroborates the underlying complexities of tracking the manifestations of violence online and will ultimately help us understand how or when online activities may result in physical violence. The recent advancements in the field of NLP have led to an increase in research focused on addressing online violence. This includes detecting hate speech, offensive language, fake news, sentiment polarity, toxicity, and online harassment in multiple languages. These efforts aim to combat the disruption of social order caused by such issues. One of the major challenges of these tasks is the lack of granularity in categorizing any content in terms of its specific context. This oversight i.e. ambiguity in the terminologies used to label content for online violence ultimately gives rise to confusion and mistrust among the researchers leading to ineffective application. On the other hand, because online violence may take many different guises it is impractical to expect complete codifications of the numerous forms of violence expressions. Thus, to strike a balance to this dilemma, we intend to focus on a specific type of online violence that is rampant in the virtual landscape of the South Asian region (Wani, 2020): communal violence. South Asia has been plagued with communal attacks and conflicts throughout the last century (Tambiah, 1990), and the digital divide along with resource scarcity is making it worse (Miklian and Hoelscher, 2017). In our study, we emphasize the Bangla-speaking regions of South Asia as a case study as it has fresh and ever-occurring scars of inter-communal disputes that often bleed to online spaces.

---

[1]Communal Violence Repository

Following the discussion above, it is evident that there is an emerging necessity to build automated systems to monitor and flag Bangla sectarian texts in the online spheres. To aid this objective, through our work we contribute the following resources:

- **Novel Theoretical Framework:** A novel extensive and representative theoretical framework for the computational analysis of communal violence in social media texts.

- **Dataset Construction Framework:** A dataset construction framework that accommodates expert opinions from social scientists, psychologists, and linguists.

- **Novel Dataset:** A novel dataset for analysis of communal violence marker from social media texts. The dataset contains 16 class expert annotations for ĩ3,000 diverse social media texts. We make the codes for our benchmarking models and the corresponding data analysis publicly available under the CC BY-SA 4.0 license.

- **Data Analysis and Benchmark Results:** Our benchmark results show that our framework and dataset are effective and reliable. This highlights the potential of computational analysis in addressing the challenges of detecting communal violence in social media.

## 2 Background

### 2.1 Contextual Relevance

Communal Violence is characterized by a sense of collective identity (MacQueen et al., 2001; Cobigo et al., 2016) and a history of tension or animosity between the groups involved. Such violence is typically driven by religion, ethnicity, language, caste, class disparity, disability, sexual orientation, cultural/political differences, or other identity markers (Brosché and Elfversson, 2012). The underlying intention may be the desire to assert dominance over another community, to retaliate for a perceived wrong, or to gain political power. The potential for communal violence is high in a region as diverse as South Asia, where numerous ethnic and religious groups live in close proximity. In recent years, there have been reports of communal violence in many parts of the region, notably Myanmar, Indonesia, and India. In Myanmar, since 2015 the Rohingya crisis has resulted in widespread violence against the Rohingya Muslim minority by the Buddhist-majority government and the military[2]. In Indonesia, there have been reports of sectarian violence between Muslims and Christians, particularly in the Maluku islands[3] which sparked early in 1999. 2020 saw a group of more than 200 people vandalizing a mosque in the village of Bhora Kalan in the north Indian state of Haryana. The group also reportedly issued threats to expel Muslim residents from the area[4]. Currently, the global refugee catastrophe fuels more communal violence due to the economic strain this puts on the host countries, and intensifies tensions between locals and asylum seekers (Böhmelt et al., 2019). The reactions to these events are essentially more spewing of communal hatred, which reaches the online spheres with even more vehemence.

In this context, Bangladesh is a particularly interesting breeding ground for communal violence due to various interconnected factors like religious affinities, political discourses, social structures, ethnicity, language, and more. Even before its birth, the region of and surrounding Bangladesh has been plagued by communal violence, as numerous communal riots from the British colonial era to the Pakistan period have shaped the values and perceptions of the people belonging to different communities (Das, 1990). Here, the media and religious/political leaders play an active role in promoting communal disparity by creating an illusory truth effect (Arendt, 2005; Hassan and Barber, 2021). The country has a history of attacks on religious and ethnic minorities[5], casually practices domestic violence on women[6], and has a prevalent discriminatory outlook towards underprivileged communities[7]. Likewise, these aspects make Bangladesh an ideal and resourceful choice for studying communal violence (Cain et al., 1979; Chowdhury, 2009; Islam and Hasan, 2021), especially in its evolved form influenced by globalization.

---
[2]Myanmar Rohingya Genocide
[3]Indonesian sectarian violence
[4]India Religious expulsion

It is important to acknowledge that the causes and expressions of communal violence are complex and multifaceted, and vary depending on the specific context and circumstances. These motivators can be both internal & external, which combine with the ubiquity of virtual accessibilities to contribute to the recent surge of intolerance in online spaces. In fact, online platforms can provide an incubator for hate speech and extremist ideologies as it provides a veil of anonymity which in turn provides security to the aggressors (Zimmerman and Ybarra, 2016). It can be used to coordinate and plan acts of violence and spread misinformation all of which can lead to further tensions and escalate the situation. On this basis, monitoring online activities and tracking the spread of hate speech can provide early warning signs of potential outbreaks of violence (Brooten, 2020), allowing authorities to take proactive measures.

## 2.2 Related Works

The task of identifying violent or toxic content in the online space has gained much popularity in the NLP domain. Previous research in the field has focused on the automated recognition of various related behaviors, including trolling (Kumar et al., 2014; Mojica de la Vega and Ng, 2018; Mihaylov et al., 2015; Cambria et al., 2010), radicalization (Agarwal and Sureka, 2017, 2015), racism (Greevy and Smeaton, 2004), flaming/insults (Nitin et al., 2012; Sax, 2016), and cyberbullying (Nitta et al., 2013; Dinakar et al., 2012; Dadvar et al., 2014, 2013). In addition, several shared tasks and academic events across multiple different languages have been organized, which focus on tackling this problem for a large amount of user-generated content. Very recently a SemEval task on identifying hate speech targeted towards women (Kirk et al., 2023) has seen participation from over a hundred different teams around the world. Similar tasks on hate speech targeted at immigrants and women also appeared in previous SemEvals (Basile et al., 2019). While the former only covered the English language, the latter task also includes a Spanish dataset. There have been several tasks on identifying offensive and toxic text spans (Zampieri et al., 2019) (English) (Pontiki et al., 2016)(Multilingual) (Wiegand et al., 2018)(German). Another vein of the task, focused on detecting overt and implicit aggression in social media contents (Kumar et al., 2018) (English/Hindi) (Álvarez-Carmona et al., 2018) (Mexican) (Tonja et al., 2022)(Spanish). The diverse perspectives in this field have led to a multitude of terminologies and understandings of the phenomenon. While this offers valuable insights, it has also created a theoretical gap in understanding their interrelationships. There is also duplication of research and a lack of focus and dataset reusability across different research strands.

---

[5]Religious Census
[6]BD women oppression
[7]BD Indigenous land-grab

## 2.3 Motivations

This work can contribute in two ways for two main reasons. Firstly, it goes beyond the primary objective of creating resources to monitor the spread and intensity of violence toward vulnerable communities. Secondly, it is relevant to the current online and social context of South Asia. Additionally, the work emphasizes the importance of categorizing the data with fine-grained labels that are meaningful from a socio-pragmatic perspective.

It is undeniable that the target labels are not mutually exclusive, making the problem both a multi-class and a multi-label classification challenge. Also, by considering the categories and expressions of communal violence orthogonally, the nuances of human interactions online can be captured more effectively compared to the recent datasets. Thus, this taxonomy not only presents a challenging machine-learning problem but also helps to establish the model's acuteness in a broader sense. Our detailed characterization of the content and the rationales behind each decision can be utilized by researchers from varying branches to outline the nature and extent of the problem. The ubiquity of the theoretical formalization can also be translated to other domains.

Regardless of being a significant conflict and communal violence-prone region[8], there hasn't been notable research in studying the unique violence motifs of the geographically Bangla-speaking region. Thus, one of our crucial motivations has been capturing the conflict and violence scenario of this particular geopolitical neighborhood. Our dataset explicitly focuses on the events and topics relevant to the landscape of this region, that received significant reciprocity from the locals. Therefore, when choosing the data collection platforms we have carefully selected the ones where the interactions from the Bangla-speaking region are more prominent. While the quantity of digital Bangla content on the internet, is ever on the rise - there are no automated ways to monitor the unhinged morbidity perpetuated through this language. Thus, the Bangla online spaces are often gold mines for documenting online violence. Consequently, we have also been driven by the fact that inadequate quality and resources remained a great obstacle in the path of developing state-of-the-art NLP frameworks for Bangla. Even though multitudes of related datasets and shared tasks (See Section 2.2) exists in various languages, we contribute the first and most extensive Bangla dataset for this particular task for the currently under-resourced Bangla NLP scene.

## 3 Proposed Taxonomy

Annotation of communal violence inciting texts is a challenging task due to the existence of divergent valid beliefs regarding the concepts. Subjectivity is a prominent challenge regarding NLP tasks such as online toxicity and harmful content detection (Jiang et al., 2021; Al Kuwatly et al., 2020; Sap et al., 2019), hate speech detection (Waseem, 2016; Salminen et al., 2019; Davani et al., 2021), stance detection (AlDayel and Magdy, 2020; Luo et al., 2020) , and sentiment analysis (Kenyon-Dean et al., 2018; Poria et al., 2023). To address this, we adopt the idea of the Prescriptive Annotation Paradigm' (Davani et al., 2021).

### 3.1 Stratification of Violent Content

While conducting research on incivility in human communication discourse, past academics operationalized a number of concepts to classify the scope and nature of incivility emanating from their study population (Sydnor, 2018; Sobieraj and Berry, 2011) . Additionally, as communal violence is a complex social phenomenon, and interconnected with other domains - it is necessary to hold a multidisciplinary research perspective (Jabareen, 2009). The current study employs a method of operationalization based on previous literature and existing settings of the research to classify violence-related categories in order to restrict the scope of the study to a feasible boundary, while making a comprehensive pipeline. The proposed taxonomy of this qualitative and quantitative research utilizes both deductive and inductive methods. The underlying intuition for this is based on the assumptions of the conceptual framework approach of categorization (Grant and Osanloo, 2014). Social science researchers often argue that the categorization process may use both approaches simultaneously i.e. creating new concepts in a deductive approach based on the principles of the inductive approach (Armat et al., 2018; Elo and Kyngäs, 2008). Therefore, We start with a deductive reasoning approach, which borrows from a previously shared task on Misogyny detection (Kirk et al., 2023). However, the motivation of the EDoS task is culturally and contextually different from our particular research setting and problem. Thus, following an inductive approach to categorization, we refine the aforementioned (EDoS framework) based on our specific context.

We developed our annotation guideline based on the taxonomy discussed in the following subsections. Our guideline contains detailed definitions, examples, exceptions, edge cases, clarification notes - as well as, establishes boundaries of inclusion and exclusion of certain demographics and defines the commune membership under specific categories. In table 7 we compile a comprehensive collection of examples for the corresponding labels in our proposed taxonomy.

---
[8]https://www.visionofhumanity.org

### 3.1.1 Identity Based Dimensionality of Communal Violence

Communal violence involves targeting victims based on their group membership, particularly religious and ethnic groups. We have kept two classes (**Religio-communal violence** and **Ethno-communal Violence**) to cover these, respectively. However, communal violence extends beyond these dominant types to include the persecution of minorities based on socio-cultural, linguistic, or nationalistic identities. To address this, our framework includes a separate category for other communal violence cases called **Nondenominational Communal Violence**. Additionally, a general violence category (**Noncommunal Violence**) is established to distinguish communal violence from other non-violent situations. Detailed definitions for the four identity-based violence dimensions are provided in the Appendix.

### 3.1.2 Identifying the Expressions of Violence

As previously discussed, online violence is a complex phenomenon that can take many different forms, ranging from threats of physical violence such as assault and homicide to psychological violence such as harassment and bullying. Based on earlier works (Besley and Persson, 2009; Simon and Greenberg, 1996; Ehrlich, 1973) in this field, we have defined four exhaustive and comprehensive categories based of severity and prevalence: i) **Repression** (Threat of harm) ii) **Prejudication** (Act of expressing unjust or unfounded opinions/false accusations) iii) **Antipathy** (Negative attitude, disgust, hatred) and iv) **Derogation**(Insultful remark, communal slur, dehumanizing/disrespectful statement). In this study, we consider these features orthogonal as multiple expressions of violence can co-occur in the same text and no objective hierarchy can be established on the severity of the expressions. A brief conceptualization of our categories is presented in Appendix Section A.0.2. We also show the different Expressions of violence and their operational definitions in table 1.

### 3.1.3 Justification for Orthogonality

While several earlier works have opted for using hierarchical categorizations of the expressions of violence - we argue that orthogonal analysis of these expressions because such hierarchies are often not aligned with the real-life samples, i.e: it is unlikely that the 'quantity' or 'severity' of violence is always going to distinctly differ from class to class. For example, the derogation category. There are many different types of derogatory comments expressing varying degrees of aggression. Thus, if we adopt a degree of severity scheme by assuming some type of hierarchical relationship among the classes, it is not guaranteed that a comment of a derogatory class will always be less severe than a comment which contains contents classified as Repression. It can sometimes be more aggressive than any other category (repressive, prejudiced, antipathy) with a higher score. Similarly, many comments in the previous categories (repressive, prejudice, and antipathy) may receive the same marks as the derogatory category for less intensity.

## 4 Dataset Construction

In this section, we present our approach to creating a high-quality annotated dataset, BASED (**B**angla **A**dverse **S**ectarian **E**xpressions **D**ataset) - focusing on reducing implicit bias and ensuring diversity in our team of annotators. To this end, we recruited eight graduate-level social science experts from diverse socio-cultural and religious backgrounds, with a particular emphasis on including members from communities affected by communal violence. We also have one Linguist with 2 years of professional experience and a psychologist with 4 years of professional experience in our expert validation team. We follow a 7-step process to collect, annotate, validate, and de-bias the data iteratively. We show the annotation pipeline in Figure 1. We discuss the dataset preparation & curation process below:

### 4.1 Collection and Processing of Samples

While a large number of related research relies entirely upon `Twitter` and `Reddit` to sample social media interactions (Vidgen and Derczynski, 2021) - neither of these platforms is very popular in Bangladesh. Thus we had to identify online spaces popularly used by Bangla content creators, i.e.: attracted a large number of user interactions. We collected the samples from the comments section of the identified virtual and social media posts. While collecting, we ensured the representation of the different identity-based and expression-based strata, as well as popular incidents of inciting communal violence in social media. We also sample some non-communal but vi-olent interactions to distinguish between communal and non-communal violence. After every iteration of data annotation and validation, we check the data and analyze the representation of the different strata and collect the next batch of samples from social media based on that. In total we have collected a dataset with $30k$ samples. Upon high-level filtering based on Bangla token presence, we retain $13k$ comments for annotation.

| Category | Vector | Definition |
|---|---|---|
| 1. Derogation | 1.1 Semantic derogation | Characterizing or describing in a derogatory manner while also producing incivil comments, vulgar language, slurs, or insults. |
| | 1.2 Emotive attacks | Dehumanizing, bullying or harassing. |
| 2. Antipathy | 2.1 Creating a distance | Differentiating, alienating, or labeling as other as well as rooting for deportation. |
| | 2.2 Hostile approach | Asking for justice from a personally biased perspective, encouraging stripping others rights, and showing strong negative sentiment such as hatred, disgust, hate speech. |
| | 2.3 Actions out of superiority complex | Demanding privileges over others because of superior feelings and offering unsolicited or patronizing advice. |
| 3. Prejudication | 3.1 Foul play | Supporting, justifying, or denying intentional or bona fide misdeeds, mistreatments, and discriminations. |
| | 3.2 Accusing and blaming | Accusing falsely, without proof, or out of emotional catharsis, and blaming the victims. |
| | 3.3 Stereotyping | Generalizing in a negative manner or stereotyping. |
| 4. Repression | 4.1 Active repression | Expressing intent, willingness, or desire to personally harm or inciting or encouraging others to harm. |
| | 4.2 Indirect repression | Asking a supreme entity such as God or a powerful entity such as a state to harm or prospecting harms where the actor may be implicit. |

Table 1: Expressions of communal violence and their operational definitions

The spelling mistakes, emoticons, and unnormalized Unicode are not fixed as we leave it as a realistic challenge associated with this task. But we ensure that the comments present in the dataset contains more than three words, doesn't contain romanized Bangla texts or HTML/XML tags, no words from other languages and is perfectly anonymized (exclusion of username, URL, email, phone numbers, etc.).

| Category | Train | Valid | Test |
|---|---|---|---|
| Religio | 727 | 234 | 260 |
| Ethno | 129 | 39 | 40 |
| Nondenominational | 32 | 11 | 19 |
| Noncommunal | 1853 | 670 | 605 |

Table 2: Distribution of Different Identity Categories Present In the Dataset

### 4.2 Annotation

We developed a novel and comprehensive annotation protocol (See Fig. 1) based on the framework presented in section 3. We constructed an expert annotation team of 8 people from a social science background, one expert linguist, and one practicing psychologist. The annotators underwent a rigorous two-day training program followed by several online clarification meetings during the annotation process. To mitigate potential biases and ambiguities in the annotated data, we enlisted the help of a psychologist with several years of experience. Additionally, a linguist coordinated the entire annotation process, intervening when necessary to resolve ambiguities and update the annotation protocols accordingly. To ensure the quality and consistency of the dataset, we chose expert annotation over crowdsourcing, to annotate 20,000 sentences.

| Dataset | Train | Valid | Test |
|---|---|---|---|
| Full | 7674 | 2558 | 2558 |
| Flagged | 2741 | 954 | 924 |
| Non-Violent | 4933 | 1604 | 1634 |
| Derogation | 1297 | 462 | 456 |
| Antipathy | 506 | 160 | 165 |
| Prejudication | 1274 | 445 | 411 |
| Repression | 308 | 105 | 104 |

Table 3: Distribution of Different Expressions of Communal and Non-Communal Violence in Training and Test Split

In short, our proposed annotation workflow is a **7-step** process described in Fig. 1 and it is carried out on small batches of the dataset. The process starts with 2-fold **annotations** for each sample. This is followed by **Conflict Resolution** stage by consulting with an expert. We then **Evaluate** the annotated data from each annotator for subjectivity utilizing the expertise of a professional psychologist. An expert board then **Validates** the final annotated data. For further probing, **Compu-** tational Analysis is carried out to detect anomalies. Under-represented & exceptional instances are identified and **Resampled** for a further layer of annotation. Finally, a **Qulitative Analysis** updates the protocol for the next batch iteration.

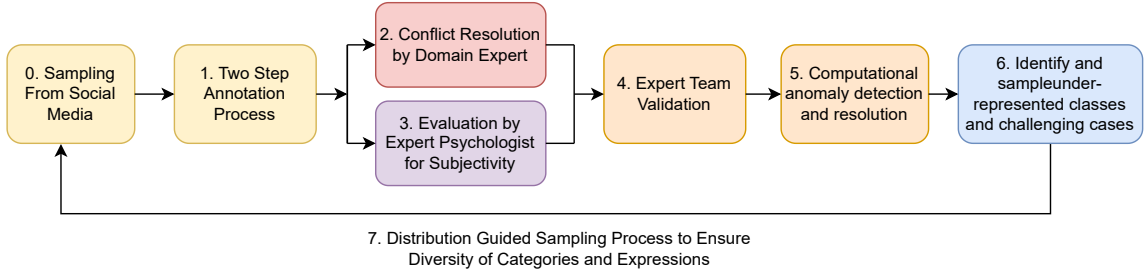

Figure 1: Annotation Pipeline

We argue that this annotation process incorporating expert insights from different relevant fields ensures good quality data for analysis and benchmarking purposes. As there are multiple checks for each data sample, we resolve conflicts and minimize subjectivity at every step, and are able to follow the protocol closely. Each annotator worked for 3 hours daily for 20 days. Throughout the process, we actively prioritized the well-being of our annotators and incorporated their feedback into our guidelines.

### 4.3 Exploratory Data Analysis

The data is split into a 60-20-20 ratio. In total, there are 18740 samples in this dataset. Each of the samples contains one/more sentences with 17 words on average. The longest sample is as many as 459 words. We have provided the distribution of the identity classes and the expressions in table 2 and table 3, respectively. From the tables, it is evident that there are substantial imbalances in the class representations. While the distribution of violent and non-violent contents is comparable, we see very low numbers of samples from the Nondenominational and Ethno communal violence categories. This imbalance in distribution reflects the prevalence of some types of online communal violence over others which can be attributed to the historical social structure in this region. The imbalance also poses a computational challenge.

As the classes are orthogonal, many sentences contain more than one marker of communal violence. To investigate their co-occurrence, we study the co-occurrence heatmap of the different expressions of Communal violence in Figure 2. The heatmap shows interesting patterns, such as, that Noncommunal violence is more often expressed through prejudication and derogation. On the other hand, the most prominent violence category in our corpus Religio-communal violence may appear quite evenly in all four forms of violence expressions. Manifesting violence through Repression is less common than the forms of expression.

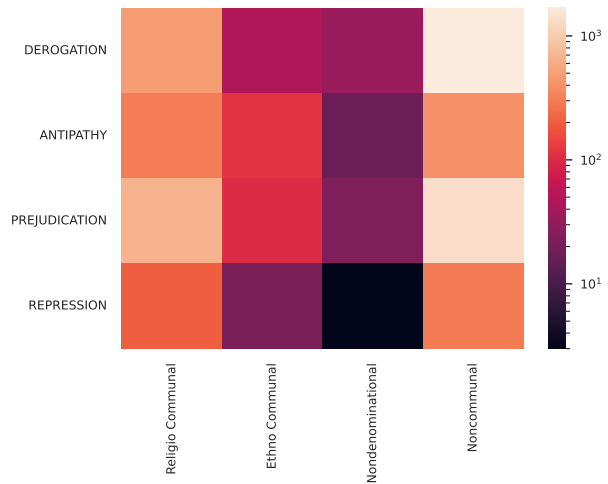

Figure 2: Heatmap of the Co-occurance of different classes

## 5 Experiments

We have designed two separate experiments distinguished by their level of granularity, to evaluate the utility of our dataset. The tasks reflect the complexity of our proposed taxonomy.

### 5.1 Experiment 1: Multilabel Classification of Communal Violence Categories

Provided a text comment from an online platform, the objective of this task is to predict which of the four Communal Violence Stratification categories the content falls into. It is possible for a content to contain semantic markers for multiple non-overlapping Communal Violence class (See section 3). This makes the task multilabeled by nature. A lack of any type of class flagging in the task implies that the sample instance does not contain violent motifs / is non-violent reaction to violence or provocation.

|  | Model | Epochs | Accuracy | Macro F1 |
|---|---|---|---|---|
| Sixteen Class | BanglaBERT | 12 | 61.5% | 18.5% |
| Four Class | BanglaBERT | 9 | 67.0% | 36.6% |

Table 4: High-level performance of baseline models

|  | Ethno-communal | Religio-communal | Nondenominational | Noncommunal |
|---|---|---|---|---|
| - | 42.9/45/44 | 46/49.2/47.5 | 0.0/0.0/0.0 | 51.6/57.2/54.2 |
| Repression | 0.0/0.0/0.0 | 53.8/19.5/28.6 | 0.0/0.0/0.0 | 47.5/32.2/38.4 |
| Prejudication | 50.0/9.1/15.3 | 30.9/33.2/32 | 0.0/0.0/0.0 | 28.9/29/28.9 |
| Antipathy | 40.0/10.6/16.7 | 22.3/3.4/5.8 | 0.0/0.0/0.0 | 22/16.1/18.6 |
| Derogation | 0.0/0.0/0.0 | 52.2/44.9/48.2 | 0.0/0.0/0.0 | 59.2/49/53.6 |

Table 5: Precision/Recall/macro-F1 Percentage of Four-class(1st row) and Sixteen-class classification using pre-trained transformer models

### 5.2 Experiment 2: Categorizing Expressions of Communal Violence towards specific demographics.

In this experiment, we try to classify the expressions of violence along with the identity-markers. As the expressions are orthogonal (See section 3.1.2), there are in total 16 classes to classify. This categorizartion can be considered as a constrained multilabel classification. The absence of any of these classes would mean an implicit non-violence class.

### 5.3 Modeling

We benchmark our dataset through our designed 4-class and 16-class classification tasks by fine-tuning a transformer-based pre-trained language model, BanglaBERT (Bhattacharjee et al., 2022). uses th Transformers are powerful for capturing long-range dependencies and global context. By fine-tuning a pretrained transformer, we can leverage their initial large-scale training and avoid the need for training the model from scratch. The models are trained using standard parameters without any targeted hyperparameter-tuning (See App. A.2). We also do not perform any normalization, sampling or other pre-processing on the training data for these baselines.

### 5.4 Benchmarking Metrics

We use the same standard evaluation metrics to evaluate our data for both granularities, and report the results for the two approaches in Table 4. For each of the label classes Precision, Recall, F1-score (See Table 4 and 5 ) and the confusion matrix is calculated independently (See App. A.2). As the experiments are multilabeled in nature - we aggregate F1 scores using a macro average scheme (Özgür et al., 2005). The macro F1-score is simply the average of F1-scores for all labels in the dataset, weighted equally regardless of the number of examples. This makes the metric particularly suitable for imbalanced datasets like ours.

From Table 4 we observe that the BERT-based approach outperforms the LSTM models by a large margin, but both fail to classify correctly for a significant number of cases. As the number of classes increase, macro-F1 performance drops for both models. The performance drop can be attributed to multiple aspects. Firstly, the corpus contains data imbalance for certain granual labels. From table 2, we know that there are only 62 Nondenominational communal violence comments. As we donot do any artificial sampling or pseudo labeling priorhand (so to reflect-the real life scenerio), this visibly impacts the evaluation metrics as well. Secondly, we only propose the models as baselines to show the applicability of the task. None of the models are explicitly tuned to achieve the best performance. Neither are the training data modified for better adaptations to the pre-trained weights. Finally, the reported scores clearly establishes the computational challenges associated with this task. Developing appropriate frameworks for this challenge would require further probing of the dataset and the contexts.

### 6 Conclusion

In this work, we presented a novel framework for analyzing communal violence texts in social media through categorization based on identities and expressions. We develop a dataset with 13k samples and present a computational analysis of the data and also benchmark model performance. We demonstrate that even current SOTA language models for Bangla are inadequate to consistently provide predictions for communal violence categories, due to the nuanced semantic and contextual challenges associated with this task. We present this as a downstream task for language model benchmarking. Also, we leave the extension of this dataset to include more target groups and introduce word/phrase-level annotation for granular analysis as future work.

## 7 Limitations

The implications of expertly annotated communal violence content are crucial for understanding and combating harmful online behavior. But it remains a difficult and complex goal requiring careful consideration of the limitations inherent to the task. Below, we briefly discuss these constraints:

**Linguistic Complexity:** The annotation task faces challenges due to the dynamic nature of the Bangla language, which evolves in response to social and cultural contexts [9]. This complexity makes it hard to accurately stratify and label communal violence content, considering linguistic nuances and subtle meaning variations. Additionally, the presence of different Bangla dialects and accents further complicates the accurate identification of violent content, particularly for annotators unfamiliar with specific dialects.

**Oversimplification:** Categorizing all violent content into four categories oversimplifies the complex linguistic situations underlying textual violence. This compression of content within categories amplifies differences and may lead to discrimination and a rigid categorical structure. Moreover, societal changes can give rise to new types of communal hate speech that do not fit into existing categories or have slightly different characteristics, challenging the fixed nature of these categories.

**Frame of Reference:** Annotating communal violence texts poses the challenge of subjectivity. Varying interpretations of hate speech among annotators can result in inconsistencies and inaccuracies in the annotated data. This subjectivity becomes pronounced when addressing subtle forms of violence like microaggressions or dog whistles, which are challenging to detect and categorize accurately. Additionally, the lack of annotators from all targeted groups limits their ability to fully understand the experiences of those subjected to online communal violence. This lack of perspective may hinder accurate categorization and the comprehension of contextual intent.

**Data Imbalance:** As a binary classification task, our dataset has a balanced distribution of violent and non-violent content. However, as we increase the granularity we find that there are substantial imbalances in the class distribution. Notably, the dataset has a low representation of Ethno-communal and Non-denominational communal categories among the violent content.

## Ethics Statement

Through our efforts, we have developed a fine-grained repository of real-life instances of online communal violence incitations. Previously we have discussed how our data and framework has certain limitations. Additionally, we also demonstrated in our analysis the particular cases where state-of-the-arts models might err in identifying labels. The combinations of these two could be leveraged by malicious actors to bypass the detection systems in the future. In a more severe case, they may be motivated to train generative models to perpetuate automated misinformation and violence in the online spheres. However, we argue that the qualitative impact and monitoring capacity of this research is much higher than these stakes. We also hope that through iterations, many of these risks can be curbed.

Another risk associated with releasing this dataset is the repeated exposure of the annotators to targeted communal violence contents. As most of the annotators fall into the targeted groups, the intensity of this task impacts their mental health. To prioritize the well-being of our annotators, we adhere to strict protocols and maintain open lines of communication through a dedicated group messaging forum. In addition, we have enlisted the expertise of a professional psychologist who is also a co-author of the paper. This ensures that annotators receive regular support for their mental health, creating a safe environment throughout the annotation process.

---

[9] Evolution of Bangla

## A Appendix

### A.0.1 Annotation Protocol

We have tried here to use language as a pragmatic model for the problem. Here we integrate context to make hierarchies in those four categories. As

Table 6: Sample Annotation Of Dataset. The annotations 1,2,3 and 4 are the four expression categories, namely Derogation, Antipathy, Prejudication and Repression.

| Sentence | Rel-Com | Eth-Com | Oth-Com | Gen-Vio |
|---|---|---|---|---|
| আপনার গঠনমূলক বক্তব্যার সরকারের নজর দেওয়া উচিৎ, অন্যথায় এই বেইমান রোহিঙ্গারই হবে বাংলাদেশের জন্য হুমকি, ধন্যবাদ। | 0 | 1,3 | 0 | 0 |
| সফিকুল দা এইসব বুদ্ধিজীবীদের থেকে সভা কিছু আশা করা বোধহয় ভুল হবে....... | 0 | 0 | 0 | 1 |
| ওরা স্বাধীনতা চাচ্ছে ওদের কে ট্রেনিং দিয়ে বার্মায় পাঠিয়ে দিন এটাই সরকারের সঠিক কাজ হবে আমি মনে করি | 0 | 4,2 | 0 | 0 |
| নায়ক থাকলে দেখা যায় তুর্কি প্রেসিডেন্ট | 0 | 0 | 0 | 0 |
| বাংলাদেশ থেকে আপনার জন্য দোয়া এবং ভালোবাসা রইল। ❌❌❌❌❌❌❌❌❌❌ | 0 | 0 | 0 | 0 |
| মুসলমান মুসলমানের ভাই,,তাই রোহিঙ্গা মুসলমানের ভাই,,এবার দেখবো ভাই ভাই যুদ্ধ,আল্লার গজব | 0 | 0 | 0 | 0 |
| মজিদ আপনাদের আল্লাহ এর পবিএ ঘর.... সেখানে যখন জঙ্গি আক্রমন করে তখন কি আল্লাহ কি করে.... | 1,2 | 0 | 0 | 0 |
| বান্দরের এখন চেক দিয়া লাগবো,,, ছেমরি বান্দর নাকি বাটা বান্দর ❌❌ | 0 | 0 | 0 | 1 |

a pragmatic language model, we have three properties here (following the definitions of (Alston, 2000):

- **Lucutionary Act:** Conventional meaning of speech
- **Illocutionary Act:** Who made the comment, and, in what context, what is the intended meaning
- **Perlocutionary Act:** Response

Based on this, we prioritize on the semantic markers first, and then context in the following manner:

- First priority: Look for semantic, literal, denotative sources and/or aspects of violence in a comment
- Second priority: Use the context to look for possible sources or aspects of violence
  - Do not forcefully assume any context of a comment if you are not sure about
  - Only take the context into consideration when you have a decent idea of what event the comment might related to be
- Flag based on targeted groups. Asking Allah to punish the government because it hanged the militants will not be in Religio-communal, but in Generic violence.
- If no semantic aspect of violence is found, then seek for a contextual approach and proceed with that.

In step 1, our approach contains a semantic filter. We have therefore divided into sixteen categories on the basis of (1) **Identity:** the different minority (or sometimes, majority) groups who are victims of communal hate speech, and (2) **Expression:** the different expressions of communal violence.

In the second step, we consider the identity of the person who made the comment and see the responses of the different community members. If the same person makes generalized comments about his community, it may not be hate speech. So its mark will be negative from step 1. Again, if such generalized comments are made in the case of different identities, then it will be considered hate speech. Then its mark will be positive. Sometime during the use of rhetoric, it is very difficult for an annotator to identify subtle uses of hate speech.ă In that case, we will see the response of the identity or community that the speech is aimed at. Suppose person A uses subtle hate speech (using metaphor or allusion) toward person B. If person B in that community makes a certain response, then we can identify it as hate speech. As we are working with single texts and not conversations, we look for semantic clues if a given text is actually a response that falls under the condition mentioned above.

### A.0.2 Categorization of Violence Expressions

The target of such expressions could be a person, a community, an entity, or an ideology. These categories encompasses both explicit and implicit expressions of violence.

**Derogation**

- Characterizing or describing in a derogatory manner
- Showing incivility
- Producing derogatory comments, vulgar language, slurs, or insults
- Dehumanizing
- Bullying or harassing

**Prejudication**

- Supporting intentional or bona fide misdeeds, mistreatments, and discriminations
- Justifying intentional or bona fide misdeeds, mistreatments, and discriminations

- Denying intentional or bona fide misdeeds, mistreatments, and discriminations
- Accusing falsely, without proof, or out of emotional catharsis
- Blaming the victims
- Doing Negative generalizations or stereotyping

**Antipathy**

- Differentiating, alienating, or labeling as other
- Rooting for deportation
- Asking for justice from a personally biased perspective
- Encouraging stripping others rights
- Showing strong negative sentiment such as hatred, disgust, hate speech
- Demanding privileges over others because of superiority complex
- Offering unsolicited or patronizing advice

**Repression**

- Expressing intent, willingness, or desire to harm
- Inciting or encouraging others to harm
- Asking a supreme entity such as God or a powerful entity such as a state to harm
- Prospecting harms for a person or a community where the actor may be implicit

The whole protocol with detailed example was shared with the annotators. [10].

## A.1

### A.1.1 Detailed Definition of Identity Based Classification of Violence

*Religio-communal Violence* comprises of statements intended to instigate violence, frequently disproportionately, towards a group with a particular religious belief, in the majority of cases the minority, converts, and non-believers. The communities of this category are the people of several religious beliefs and different practices in their spirituality including muslims, hindus, christians, ahmadias, sh'ites among others, because these groups of people often find themselves in a muddy confrontation, both online and offline, with each other in regard to their religion or get dehumanized by the believers of major faith. (Rajarajan, 2007)

*Ethno-communal Violence* refers to any form of violence that is directed against individuals or groups based on their ethnic or communal identity. The affected groups include indigenous people, settlers such as Bengalis and Biharis, Rohingyas, and other ethnic groups. Indigenous people, for instance, are often targeted because they are seen as a threat to the dominant group's land ownership or access to natural resources. Settlers, on the other hand, may face violence due to their perceived intrusion into the lands of the native population. Biharis and Rohingyas are often targeted due to their refugee status or minority status in the country. (Gubler et al., 2016)

*Non-denominational Communal Violence* includes additional types of communal violence, such as violence aimed at persons from another linguistic identity (Engineer, 1994) (but not necessarily another ethnic group), district because of their geographical(Toft, 2010; Döring, 2020) and cultural differences, different languages because of their funny or kitsch perceptions to others, people with different cultural adaptations such as Bauls.

*Non-communal Violence* includes any other violence apart from these three specific categories. We keep this category to ensure that we dont miss out a bigger portion of data on violence as there are lots of different non-communal violent events that occur frequently. However, for the feasibility of the study, we only added violent incidents on an individual level, misogynistic appearance of violence(Kirk et al., 2023), violence against the LGBTQ community(Casey et al., 2019), the state (collectively, or specific state agents) (Jacobs and Carmichael, 2002) or the private sector (Phillips, 2016).

### A.1.2 Detailed Definitions for the expression of Communal Violence

*Derogation* refers to any form of communication or expression that is intended to belittle, demean, insult or harm a specific target or group of individuals. This can include the use of derogatory expressions, communal slurs, or other insulting language, as well as describing or characterizing a target in a manner that is disrespectful or uncivil. Derogatory behavior can also involve dehumanizing a target, treating them as if they are not worthy of respect or consideration, or engaging in bullying and harassment aimed at causing emotional or psychological harm.

---

[10]Protocol

| Communal Violence Class | Violence Expression | Example (Bangla) | Example (English) |
|---|---|---|---|
| Religio-communal Violence | Derogation | জেই ধর্ম সাধুরা উলঙ্গ হয়ে ঘোরে সেই ধর্মের মানুষেরা হিজাবের মর্মতা কি করে বুঝবে একটা উলঙ্গ ও নিক্রিষ্টা ধর্ম হচ্ছে হিন্দু ধর্ম রামের শিক্ষা আর কতো ভালো হবে ছি ছি | The religion (Hinduism) where the monks move about naked, what would its followers understand about Hijab? |
| | Antipathy | তদের সময় গনিয়ে আসছে দেশ ছারার সবকটাকে মালুরদেশ ভারত পাটানো হবে আর কটাদিন সবুরকর | It's almost time! All the Malauns (Slur for Hindu folks) will be deported to India. You wait! |
| | Prejudication | লোকগুলো কে মেরে ফেলতে হবে কেনো!মাদ্রাসার বেশির ভাগ হুজুর সমকামী! এদেরকে তো মেরে ফেলার কথা কেউ বলে না! | Why must we kill these people? All the Hujurs (Islamic teachers) in madrasas are homosexual! Why is nobody asking to kill them? |
| | Repression | এটা বাংলাদেশে হলে গুলি করে মেরে ফেলত, বাংলাদেশে উগ্রবাদী হিন্দুদের সোজা করা উচিত, যাতে বাংলাদেশে ঝামেলা করতে না পারে। | If it was in Bangladesh, they would have been shot dead. It is imperative to straighten up the extremist Hindus in Bangladesh, so they can't cause any troubles/ |
| Ethno-communal Violence | Derogation | এই রোহিঙ্গা এখন রাজনৈতিক, অর্থনৈতিক সমাজিক ও সাংস্কৃতিক উভয় ক্ষেত্রেই গলার হাড্ডি হয়ে দেখা দিয়েছে!! এবার বুঝবে এই রোহিঙ্গা কতো প্রিয় জিনিস!! | These Rohingya are now political, economic, moral, social and cultural burden. Now you'll understand how adored the Rohingyas are! |
| | Antipathy | রোহিঙ্গাদের দেশে রোহিঙ্গাদের পাঠানো হোক | Deport the Rohigyas back to their own country. |
| | Prejudication | এমন অনেক রুহিঙ্গা ও স্থানীয় ব্যবসায়ী ওখানে আছে যারা প্রতিনিয়ত বড় বড় ইয়াবা চালান আনে। | There are many Rohingyas and local traders who regularly bring large consignments of yaba (drugs). |
| | Repression | এই রোহিঙ্গার স্বীকারোক্তি নিয়ে যত দ্রুত সম্ভব নাফ নদীতে চুবিয়ে মারেন। | Get a confession out of this Rohingya, then immediately drown him in Naaf river. |
| Nondenominative Violence | Derogation | নোয়াখাইল্লা বাটপারের পাল্লায় পরসিলাম | I fell into the clutches of a Noakhailla (region in Bangladesh) swindler |
| | Antipathy | সিলেটিভাষীদের দেশ থেকে বের করে দাওয়া হোক। | Deport the Sylheti Speakers. |
| | Prejudication | ব্রাহ্মণবাড়িয়া লিখতে হবে!? ব্রাহ্মণবাড়িয়া শব্দটি হিন্দুদের, আর বি-বাড়িয়া শব্দটি হিন্দু হিন্দু লাগেনা হিম?!! | So I have to write BrahmanBaria (region in Bangladesh)? Is it because writing B.Bariya doesn't sound Hindu enough? |
| | Repression | হাত পায়ের মায়া থাকলে বাউনবাইরা লইয়া কিছু কইয়েন না | If you care about your limbs, don't try to slander BaunBaria (region in Bangladesh) |
| Generic Violence | Derogation | যারা হা হা রিয়েক্ট দিছে এদের জন্ম সমস্যা আছে | Those reacting 'Haha' have dubious paternity record. |
| | Antipathy | যারা সমাজে সমকামিতা প্রচার ও প্রসার করতে চায়, সরকারেরইতো উচিত ছিল তাদের আইনের আয়তায় আনার। | Those who want to promote and spread homosexuality in the society, the government should have brought them under the law. |
| | Prejudication | পুলিশের তো এমন লোকেরই দরকার।কারন তারাই ইয়াবার বড় কাষ্টমার। | The police need such people because they are Yaba's (drug) big customers. |
| | Repression | ইসরাইল সন্ত্রাসী গোষ্ঠী আল্লাহ আপনি এদের ধংস করে দিন। | Israel is terrorist group God destroy them. |

Table 7: Examples of the Expressions of Violence across our defined Violence Dimensions

*Antipathy* is a term used to describe a set of negative attitudes and behaviors towards individuals or groups, which can include expressions of disgust, hatred, or other strong negative emotions. These attitudes often involve differentiating or alienating others based on communal identity, and may involve offering unsolicited or patronizing advice, demanding unjust privileges due to a superiority complex, or advocating for actions such as stripping rights or deportation. Antipathy can also refer to any form of internalized hatred that is not expressed through slurs or other overt means.

*Prejudication* refers to the act of forming unjust or unfounded opinions, attitudes, or beliefs about a particular group of people, often based on stereotypes or negative generalizations. This can lead to the support or justification of intentional or unintentional harm or discrimination towards that group, as well as blaming the victims for their own mistreatment. It may also involve making false accusations against individuals without proper evidence or as a result of emotional catharsis.

*Repression* refers to any action or behavior that involves expressing an intent, willingness, or desire to cause harm or asking a supreme or powerful entity to cause harm. It can also include inciting or encouraging others to harm, promoting violent actions or harm towards a person or community, where the actor may be implicit, or championing communal supremacy. Repression is a negative and destructive behavior that can have serious consequences for individuals and society as a whole.

### A.1.3 Data Statistics & Vizualization

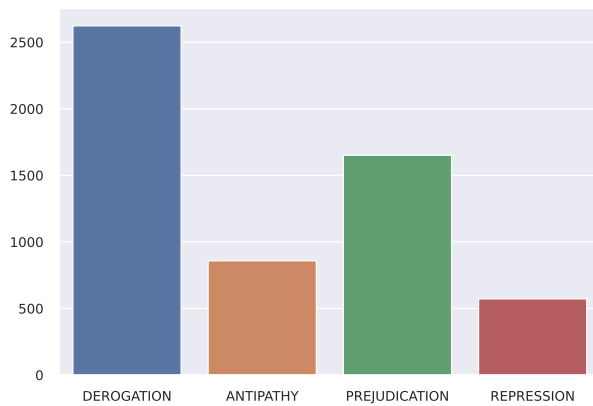

Figure 3: Distribution of Expressions of Violence

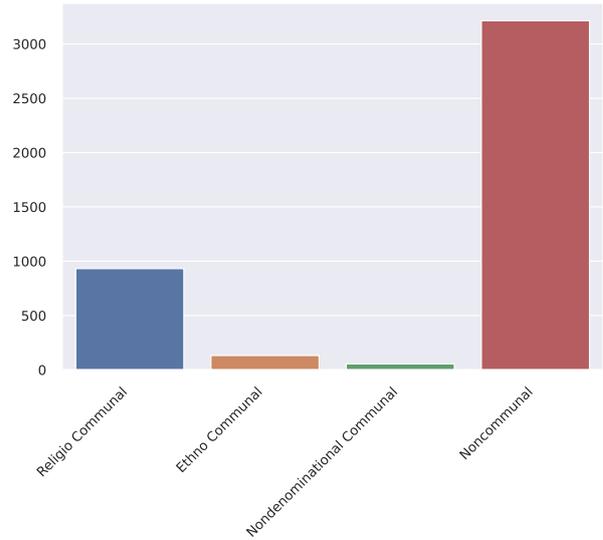

Figure 4: Distribution of Violence Targets

### A.2 Model Building & Benchmarking

| Hyperparameters | Values |
|---|---|
| Weights | banglabert |
| Optimizer | Adam |
| LR Scheduler | Linear |
| Learning Rate | 2e-5 |
| Training Batch | 8 |
| Weight Decay | 0.01 |
| Early Stopping Patience | 5 (threshold 0.01) |
| Max Epoch | 30 |

Table 8: List of Hyperparameters used to train Transformer Classification model

| | Ethno-communal | Religio-communal | Nondenominational | Noncommunal |
|---|---|---|---|---|
| - | 18/2494/24/22 | 128/2148/150/132 | 0/2539/0/19 | 346/1628/325/259 |
| Repression | 0/2551/0/7 | 7/2517/5/29 | 0/2556/0/2 | 19/2479/20/40 |
| Prejudication | 1/2534/2/21 | 49/2297/110/102 | 0/2555/0/3 | 68/2159/164/167 |
| Antipathy | 2/2536/3/17 | 2/2491/7/58 | 0/2553/0/5 | 113/2432/45/68 |
| Derogation | 0/2549/0/9 | 48/2409/42/59 | 0/2547/0/11 | 159/2120/109/170 |

Table 9: True-positive/True-negative/False-positive/False-negative scores of Four-class(1st row) and Sixteen-class classification using pre-trained transformer models

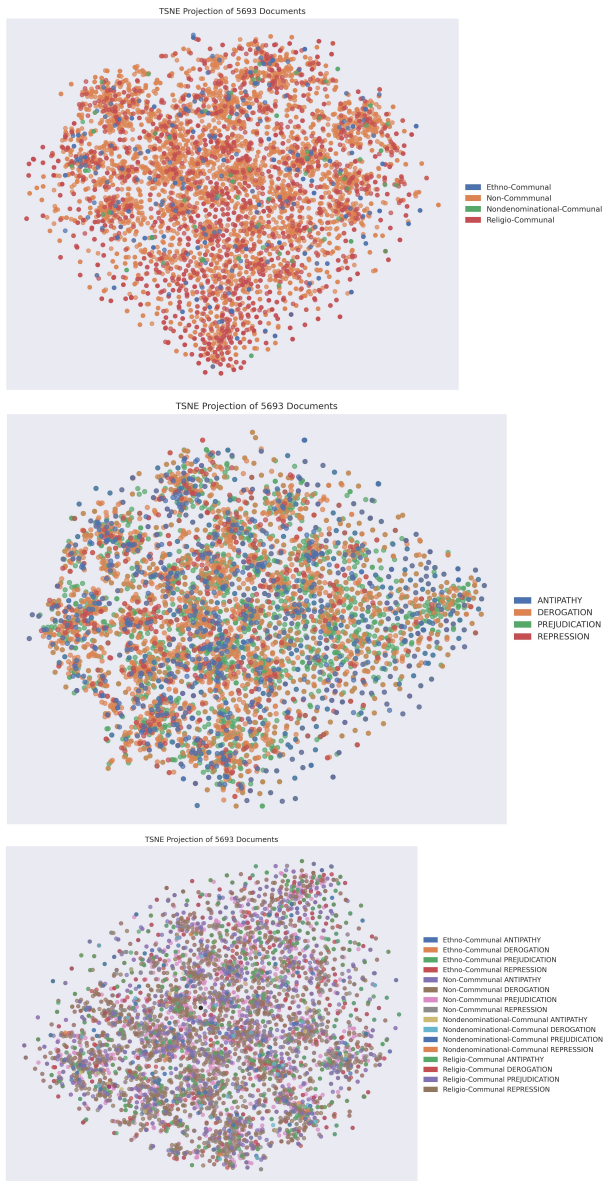

Figure 5: 2D t-SNE plot for different categories and expression of communal violence. The embeddings were calculated using BERT.